\documentclass[sigconf]{acmart}

\usepackage{url}
\usepackage{graphicx}
\usepackage{amsfonts}
\usepackage{bm}
\usepackage{algorithm}
\usepackage{algorithmicx}
\usepackage{algpseudocode}
\usepackage{amsmath}
\usepackage{mathrsfs}
\usepackage{float}
\usepackage{multirow}
\usepackage{color}

\floatname{algorithm}{Algorithm}

\newcommand{\norm}[1]{\left\lVert#1\right\rVert}

\def\BibTeX{{\rm B\kern-.05em{\sc i\kern-.025em b}\kern-.08emT\kern-.1667em\lower.7ex\hbox{E}\kern-.125emX}}

\setcopyright{acmcopyright}

\begin{document}

\copyrightyear{2019} 
\acmYear{2019} 
\acmConference[CIKM '19]{The 28th ACM International Conference on Information and Knowledge Management}{November 3--7, 2019}{Beijing, China}
\acmBooktitle{The 28th ACM International Conference on Information and Knowledge Management (CIKM '19), November 3--7, 2019, Beijing, China}
\acmPrice{15.00}
\acmDOI{10.1145/3357384.3358072}
\acmISBN{978-1-4503-6976-3/19/11}
\fancyhead{}

	\title{Interpretable MTL from Heterogeneous Domains \\ using Boosted Tree}
	    \author{Ya-Lin Zhang}
	    \affiliation{
	    	\institution{Ant Financial Services Group, China}
	    }
	    \email{lyn.zyl@antfin.com}
	    \author{Longfei Li}
	    \authornote{Both authors contributed equally to this paper.}
	    \affiliation{
	    	\institution{Ant Financial Services Group, China}
	    }
	    \email{longyao.llf@antfin.com}
	
	\begin{abstract}
		
	Multi-task learning (MTL) aims at improving the generalization performance of several related tasks by leveraging useful information contained in them. However, in industrial scenarios, interpretability is always demanded, and the data of different tasks may be in heterogeneous domains, making the existing methods unsuitable or unsatisfactory. In this paper, following the philosophy of boosted tree, we proposed a two-stage method. In stage one, a common model is built to learn the commonalities using the common features of all instances. Different from the training of conventional boosted tree model, we proposed a regularization strategy and an early-stopping mechanism to optimize the multi-task learning process. In stage two, started by fitting the residual error of the common model, a specific model is constructed with the task-specific instances to further boost the performance. Experiments on both benchmark and real-world datasets validate the effectiveness of the proposed method. What's more, interpretability can be naturally obtained from the tree based method, satisfying the industrial needs.
	
	\end{abstract}
	
%
	
	
	\maketitle              

	\section{INTRODUCTION}
	
	Multi-task learning (MTL) aims at improving the performance of several related tasks by leveraging useful information contained in them, and it has been shown empirically and theoretically that MTL can significantly improve the performance. During past few decades, many methods, such as low-rank approach \cite{DBLP:conf/icml/ChenTLY09}, task clustering approach \cite{DBLP:conf/icml/XueDC07}, have been proposed, and deep learning based methods \cite{DBLP:journals/ijcv/LiLC15} are widely explored.
	
	For industrial scenarios, a reasonable explanation is always widely needed for business requirement, making the widely-used deep learning methods unsatisfactory due to the poor interpretability. What's more, most approaches are focused on homogeneous setting, while the data of different tasks may be in heterogeneous domains in industrial applications. 
	
	Boosted tree model has been shown to be one of the best candidates for both competition and real industrial tasks, due to its superior performance~\cite{DBLP:conf/kdd/ChenG16,DBLP:journals/tist/ZhangZZFLLLZCLQ19}. What's more, the great interpretability has made it a better choice than deep learning methods in industrial tasks. However, they are mainly applied in the conventional tasks, while in settings like the multi-task learning problem, this kind of model has not been further explored.
	
	In this paper, we propose a boosted tree based algorithm for MTL, which can be divided into to stages.
	Concretely, in stage one, all instances with the common features are collected to build a common model to learn the commonalities. 
	One concern is that the tasks with a larger sample size may dominate the construction of trees, making the obtained model profitless for the tasks with smaller sample size.
	A regularization strategy is designed to alleviate this problem.
	Besides, an early stopping mechanism is proposed to get the proper round of each task in the common model.
	In stage two, started by fitting the residual error, a specific model is constructed to further boost the performance. 
	We validate the proposed method on both the benchmark datasets and real-world tasks. The results demonstrate that the proposed method can significantly improve the performance.
	
	The remainder of the paper is as follows. Section \ref{prob_state} states the studied problem, section \ref{method} presents the method. In section \ref{exps}, experiments are shown, and section \ref{con} concludes the paper.
	
	\section{Problem Statement}
	\label{prob_state}
	Given $m$ learning tasks $\{\mathcal{T}_i\}_{i=1}^m$ where all of them or a subset are related, the goal of multi-task learning (MTL) is to improve the performance of each task $\mathcal{T}_i$ by using the data of all tasks. 
	A task $\mathcal{T}_i$ consists of a training dataset $\mathcal{D}_i$ with $n_i$ samples, i.e., $\mathcal{D}_i=\{(\bm{x}^i_j,y^i_j)\}_{j=1}^{n_i}$, where $\bm{x}^i_j \in \mathbb{R}^{d_i}$ is the $j$-th training sample of task $\mathcal{T}_i$, and $y^i_j \in \mathbb{R}$ is its corresponding label. 
	Different to most of the MTL studies, we consider a partially heterogeneous setting, i.e., $d_i$ can be unequal to $d_j$ for $i \ne j$, while an overlap feature space can be found among all of the tasks. 
	Assume that the first $d_0$ features are the overlap features for all samples, then each sample can be expressed as  $\bm{x}^i_j=\{x^i_{j1},\dots,x^i_{jd_0},\dots,x^i_{jd_i}\}$, where the first $d_0$ features are the common features shared by all these tasks with the same meaning, and the rest are task-specific features.
	The label space of all these tasks is consistent with the same meaning in our study.
	
	\section{MULTI-TASK Boosted Tree}
	\label{method}
	\subsection{The Whole Framework}
	
\begin{figure}[htbp]
		\centering
		\includegraphics[width=8.5cm,trim={0.05cm 1.5cm 0.9cm 0.5cm}, clip]{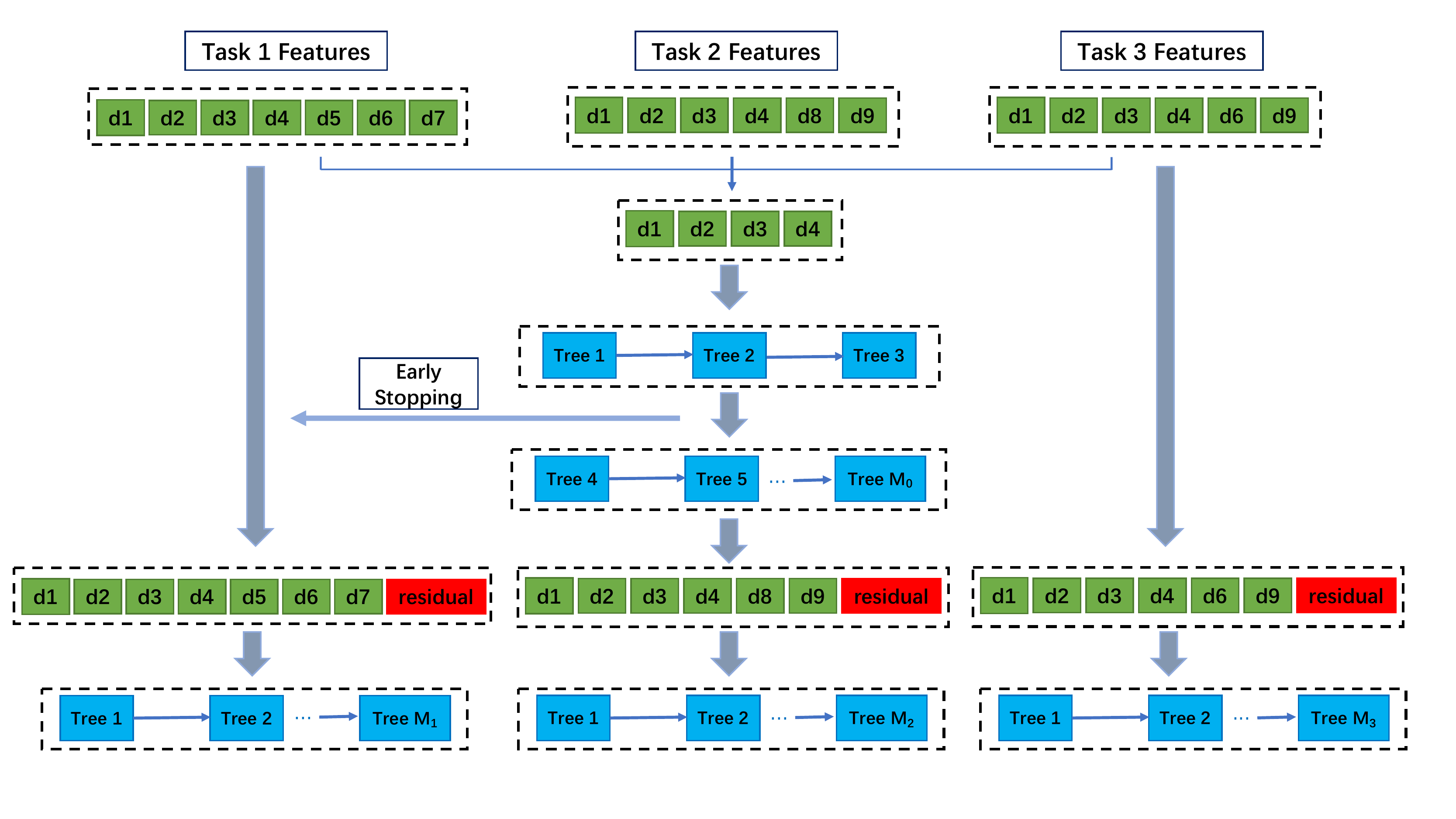}
		\caption{\small{The overall framework of the proposed method: three tasks are included, with four overlap features, which are used in the training of common model. Task 1 is early stopped after round 3 and quits from the common model.}}
		\label{overall_framework}
	\end{figure}

	Suppose we have $m$ tasks, and the data of different tasks are located in partially heterogeneous domains, 
	i.e., $\bm{x}^1_j \in \mathbb{R}^{d_1}$, $\bm{x}^2_j \in \mathbb{R}^{d_2}$, \dots $\bm{x}^m_j \in \mathbb{R}^{d_m}$, and $d_i$ can be different from $d_j$, 
	while a $d_0$-dimensional overlap feature space exists among all tasks.    
	Figure \ref{overall_framework} shows an example of the whole process of the proposed method, with three tasks and four-dimensional overlap features.
	In the first stage, we extract the overlap features and collect all instances together to build a common boosted tree model so that the commonality can be learned. In the second stage, a specific boosted tree model is built to learn the specialty and further boost the performance.
	
	However, there are some problems that need to be addressed:
	\begin{enumerate}
		\item The data size of different tasks can be severely different, and the building process of the tree may be easily dominated by some task, which means that the constructed tree in the common part may be pretty profitless for the task with smaller sample size.
		\item The construction of the trees in the common model may be unbeneficial or even harmful for some task after some rounds, which means that a mechanism is needed to find the proper round so that a task can quit from the common model training process if necessary.
		\item The training of the second stage should take the information of the first stage into consideration, so that the obtained model can be more effective, instead of simply combining two boosted tree models when predicting for each task.
	\end{enumerate}
	
	To handle these problems, a regularization strategy is proposed for the construction of each tree to alleviate the domination problem, and an early stopping strategy is designed so that a task can quit the common process if further training will not improve its performance. Besides, the specific models are started by fitting the residual of the common model. We will explain them in detail below. 
	
	\subsection{Regularization}
	In boosted tree, trees are built sequentially, and the prediction $\widehat{y_i}$ of an instance $\bm{x}_i$ from a boosted tree model with $K$ trees can be calculated as 
	\begin{equation}
	\label{prediction}
	\small
	\widehat{y_i}=\sum_{k=1}^{K}f_{k}(\bm{x}_i), f_k\in \mathcal{F} \,,
	\end{equation}
	where $\mathcal{F}=\{f(\bm{x})=w_{q(\bm{x})}\}$ is the space of regression trees. $q(\bm{x})$ maps the instance $x$ to the leaf index, and $w_{q(\bm{x})}$ is the related weight in this leaf.
	Let $\widehat{y_i}^{(r)}$ denotes the prediction of the $i$-th instance at the $r$-th iteration, then $f_r$ is built to minimize the following objective,
	\begin{equation}
	\label{objective_ft}
	\small
	\mathcal{L}^{(r)}= \sum_{i=1}^{n}l(y_i,\widehat{y_i}^{(r-1)} + f_r(\bm{x}_i)) + \Omega(f_r) \,,
	\end{equation}
	where $l$ is a differentiable convex loss function, $\Omega(f)=\gamma T + \frac{1}{2}\lambda \norm{w}^2$ is used to penalize the model complexity, $n$ is the sample number. As described in \cite{DBLP:conf/kdd/ChenG16}, with second-order approximation employed, and the constant terms removed, the objective can be simplified as,
	\begin{equation}
	\label{objective_simp}
	\small
	\mathcal{L}^{(r)}= \sum_{i=1}^{n} [g_if_r(\bm{x}_i) + \frac{1}{2}h_if_r^2(\bm{x}_i)] + \Omega(f_r) \,,
	\end{equation}
	where $g_i=\partial_{\widehat{y}_i^{(r-1)}}l(y_i,\widehat{y}_i^{(r-1)})$ and $h_i=\partial^2_{\widehat{y}_i^{(r-1)}}l(y_i,\widehat{y}_i^{(r-1)})$ are the first and second order gradient on the loss function.
	When splitting a node, a split point is selected. Let $I_L$ and $I_R$ denote the instance sets of left and right child nodes, and $I = I_L \cup I_R$ is the instance set in current node, then the loss reduction score is showed in Eq.\ref{loss_reduction}, the split point with the highest score will be selected.
	\begin{equation}
	\label{loss_reduction}
	\small
	s = \mathcal{L}_{split} = \frac{1}{2} \left[ \frac{(\sum_{i \in I_L} g_i)^2}{\sum_{i \in I_L} h_i + \lambda} +  \frac{(\sum_{i \in I_R} g_i)^2}{\sum_{i \in I_R} h_i + \lambda} - \frac{(\sum_{i \in I} g_i)^2}{\sum_{i \in I} h_i + \lambda} \right] - \gamma
	\end{equation}
	
	However, in the multi-task scenario, the whole score calculated as above may be easily dominated by the task with huge sample size, and the task with smaller sample size may be ignored, thus the chosen split point may be unbeneficial for these tasks. 
	Concretely, for each task $t$, the loss reduction score $s_t$ can be calculated by only considering the task-specific instances in Eq.\ref{loss_reduction}. Thus, if the sample size of different task is quite imbalanced, the total score $s$ may be easily dominated by the majority task, and the selected split point may be related to a small $s_t$ for the minority task, meaning that this split is unhelpful for this task. 
	
	To alleviate the aforementioned problem, an additional regularization term is proposed when calculating the best split points, so that a new tree is trained to benefit most of the tasks. 
	The regularized split finding algorithm is shown in  Algorithm ~\ref{algo_split}, where $I$ is the instance set of the current node, and $I^t$ is the instance set that belongs to task $t$. $s^t$ is the loss reduction score of task $t$. 
	Function $sorted(I, by\, \bm{x}_{jd})$ will sort the instances according to dimension $d$, $task(\bm{x}_j)$ will return the belonging task index of $\bm{x}_j$, and $S_r(s^1,\dots,s^T,s)$ will return the regularized score with the regularization term based on the original loss reduction score for each task and the whole loss reduction. In this paper, we propose two strategies, i.e., entropy-based and variance-based method, which will be explained in details.
	
	\begin{algorithm}[htbp]
		\small
		\caption{\small{Regularized Split Finding Algorithm}}
		\label{algo_split}
		\begin{algorithmic}[1]
			\Require $I$, instance set; $D$, feature dimension; $T$, task number
			\State $S_m=-\infty$
			\State $G = \sum_{i \in I} g_i, H = \sum_{i \in I} h_i$
			\For{$t=1$ to {T}}
			\State $G^{t}=\sum_{i\epsilon I^{t}}g_{i}, H^{t}=\sum_{i\epsilon I^{t}}h_{i}$
			\EndFor
			\For{$d=1$ to $D$}
			\State $G_L=0, H_L=0$
			\For{$t=1$ to {T}}
			\State $G^{t}_L=0, G^{t}_R=G^{t}-G^t_L$
			\State $H^{t}_L=0, H^{t}_R=H^t - H^t_L$
			\State $s^t=\frac{{(G^t_L)}^2}{H^t_L+\lambda} + \frac{{(G^t_R)}^2}{H^t_R+\lambda} - \frac{{(G^t)}^2}{H^t+\lambda}$ 
			\EndFor
			\For{$j$ in $sorted(I, by\, \bm{x}_{jd})$}
			\State $G_L=G_L + g_j, H_L = H_L + h_j$
			\State $G_R=G-G_L, H_R=H-H_L$
			\State $s=\frac{G_{L}^2}{H_L+\lambda} + \frac{G_{R}^2}{H_R+\lambda} - \frac{G^2}{H+\lambda}$
			\State $\hat t=task(\bm{x}_j)$
			\State $G_L^{\hat t}=G_L^{\hat t} + g_j, H_L^{\hat t} = H_L^{\hat t} + h_j$
			\State $G_R^{\hat t}=G^{\hat t}-G_L^{\hat t}, H_R^{\hat t}=H^{\hat t}-H_L^{\hat t}$
			\State $s^{\hat t}=\frac{{(G^{\hat t}_L)}^2}{H^{\hat t}_L+\lambda} + \frac{{(G^{\hat t}_R)}^2}{H^{\hat t}_R+\lambda} - \frac{{(G^{\hat t})}^2}{H^{\hat t}+\lambda}$ 
			\State $S_m = \max(S_m, S_r(s^1,\dots,s^T, s))$
			\EndFor
			\EndFor
			\Ensure Split with $S_m$
		\end{algorithmic}
	\end{algorithm}
	
	\subsubsection{\textbf{Entropy-based method}}
	First, ReLU function is applied to each $s^t$ so that it will be non-negative. We hope the loss reduction score for all tasks to be as much balance as possible. Similar to the calculation of entropy, the $S_r$ can be obtained as below,
	\begin{equation}
	\small
	\label{entropy_based}
	\begin{aligned}
	& S_r  = (-\sum_{t=1}^{T}P_{t}logP_{t}) \cdot s, \,  where \, P_{t}   =\frac{ReLU(s^t)}{\sum_{t=1}^{T}ReLU(s^t)}
	\end{aligned}
	\end{equation}
	$P_t$ can be regarded as the probability that the $t$-th task dominates the split procedure.
	
	\subsubsection{\textbf{Variance-based method}}
	The second strategy is based on the calculation of the variance of all the loss reduction scores. The details are shown in Eq.~\ref{variance_based}, in which $\beta$ is a parameter to control the influence of the variance.
	\begin{equation}
	\small
	\label{variance_based}
	\begin{aligned}
	S_r = s -\beta v, \,  where \, \bar{s} =\frac{1}{T}\sum_{t=1}^T s^t,  v =\frac{1}{T-1}\sum_{t=1}^T(s^t-\bar{s})^2
	\end{aligned}
	\end{equation}
	\subsection{Early Stopping}
	In the training process of the common model, a max iteration round is set, while it may not be the proper parameter for all tasks.
	After some rounds of training, the benefit of some tasks may not further increase or even decrease. 
	For example, the max round can be set to 100 for the whole process, but for a specific task $t$, the performance will not further improve after 10 rounds of training, according to the evaluation on a validation set. Then, this task will be removed from the subsequent training process of the common model, and the training of its specific model will start.
	An early stopping strategy is introduced by leaving out a validation set and evaluating the metric after each round. If the performance is not improved in some  consecutive rounds, the task and its data will quit from the training of the common model. If we denote the common model as $\mathcal{M}^0$, and the quit round of task $t$ as $r^t$, then the common part of task $t$ is $\mathcal{M}^0_{r^t}=\sum_{k=1}^{r^t}f_{k}$.
	
	\subsection{Specific Model Training}
	After the training of the common model, each task will get into the training of its specific model with the task-specific instances. 
	However, if we simply start the specific model by fitting the original label, it may be not natural when we combine the common model part and the specific model part. 
	With the common part of the model, the residual can be calculated, i.e., the first order and second order gradient can be obtained. We proposed to train the task-specific model by starting from the fitting of the residual information from the common model part. 
	What's more, in the common part, only the overlap features are used, while in the specific model training process, the whole features for this task are used, so that the task specialty will be addressed.
	
	\subsection{Prediction}
	The final prediction is a combination of the common model part and the specific model part. We denote $\mathcal{M}^0$ as the common model, with the quit round of task $t$ as $r^t$, and denote $\mathcal{M}^t$ as the specific model with $R^t$ rounds in all.
	Then the final prediction of an instance $\bm{x}_i$ from task $t$ will be obtained by,
	\begin{equation}
	\small
	\label{prediction_all}
	\widehat{y_i}=\sum_{k=1}^{r^t}f^0_{k}(\bm{x}_i) + \sum_{k=1}^{R^t}f^t_{k}(\bm{x}_i),  f^0_k\in \mathcal{F},f^t_k\in \mathcal{F},
	\end{equation}
	where $f^0_{k}$ denotes the tree in $\mathcal{M}^0$, and $f^t_{k}$ denotes the tree in $\mathcal{M}^t$.
	\begin{table*}
		\centering
		\caption{\small{The results of the experiments.}}
		\scalebox{1}{
			\begin{tabular}{|c|c|c|c|c|c|c|c|c|c|}
				\hline
				Dataset & Task & Measure &LASSO&RMTFL&DIRTY&GBT&IBT&VMTBT&EMTBT \\ \hline
				school&ALL&RMSE&11.21 &10.46&10.41&11.19&10.04&\textbf{8.99}&9.01 \\ \hline
				\multirow{2}*{Scene1} & task1   & AUC&0.824&0.627&0.834&0.978&0.980&0.983&\textbf{0.986} \\ \cline{2-10}
				~ & task2   & AUC&0.617&0.590&0.652&0.971&0.951&0.974&\textbf{0.976} \\ \hline
				\multirow{2}*{Scene2} & task1   & AUC &0.693&0.667&0.694&0.906&0.926&\textbf{0.942}&0.940 \\ \cline{2-10}
				~ & task2   & AUC&0.599 &0.681&0.602&0.760&0.776&0.788&\textbf{0.791} \\ \hline
			\end{tabular}
		}
		\label{exp_res}
	\end{table*}
	
	\section{EXPERIMENTS}
	\label{exps}
	
	\subsection{Datasets}
	We evaluate the performance on both public and real-world datasets.
	The first dataset comes from the Inner London Education Authority\footnote{\url{https://ttic.uchicago.edu/~argyriou/code/index.html}}, which consists of examination records of 15362 students from 139 secondary schools (which can be regarded as 139 tasks). The goal is to predict the scores of the students from different schools. 
	The other two datasets come from Ant Financial, both are about fraud detection, and each dataset is with two tasks. The goal is to find out which transaction is a fraud one. The details are explained as below and shown in Table \ref{dataset}.
	\begin{enumerate}
		\item \textbf{Scene1}: task1 is the transactions between users, while the task2 is the transactions between a user and a commercial organization, the feature space of these two tasks are not the same, with 81 common features.
		\item \textbf{Scene2}: Samples in task1 and task2 are the transactions of users who have low and normal loan amount, respectively. The two tasks are in same feature space.
	\end{enumerate}
	
	\begin{table}[htbp]
		\centering
		\caption{\small{Details of the fraud detection datasets.}}
		\scalebox{0.9}{
			\begin{tabular}{|c|c|c|c|c|c|}
				\hline
				Scene & Task & Negative & Positive  & Dimension & Overlap \\ \hline
				\multirow{2}*{Scene1} & task1   & 77k             & 2296    &     81  & \multirow{2}*{81} \\ 
				\cline{2-5} 
				~& task2  & 25k            & 989    &     202  & ~\\ \hline
				\multirow{2}*{Scene2} & task1   & 187k             & 4709         & 44    & \multirow{2}*{44} \\
				\cline{2-5}
				~ & task2   & 968k             & 11850          & 44   & ~ \\ \hline
			\end{tabular}
		}
		\label{dataset}
	\end{table}

	\subsection{Experiments Setup}
	Since the interpretability is demanded in our problem, we compare the proposed method with several state-of-the-art methods, and two boosted tree based methods, which are good at interpretability.
	The details are as below:
	\begin{enumerate}
		\item \textbf{LASSO}: This  method learns a sparse prediction model for each task independently.
		\item \textbf{Robust Multi-Task Feature Learning algorithm (rMTFL) } \cite{DBLP:conf/kdd/GongYZ12}: This method simultaneously captures a common feature sets among relevant tasks and identifies outlier tasks.
		\item \textbf{DIRTY} \cite{DBLP:conf/nips/JalaliRSR10}: This method searches for a parameter matrix that can be decomposed into a row-sparse matrix and an element-wise sparse matrix.
		\item \textbf{Independent Boosted Tree(IBT)}:  This method learns a boosted tree model for each task with its samples.
		\item \textbf{Global Boosted Tree(GBT)}: This method learns a global boosted tree model for all tasks with all samples.
	\end{enumerate}
	We use VMTBT, EMTBT to denote the proposed variance and entropy based multi task boosted tree, respectively. 
	For each method, grid search is used to find the best parameters, and the datasets are divided into two parts, i.e., 80\% for training, and 20\% for test. The average of 10-time experiments are calculated as the reported results. Note that the GBT model and other compared MTL methods cannot deal with the heterogeneous tasks, so the non-overlapped features are padding with 0 separately for each task. 
	The root mean squared error (RMSE) and the area under curve (AUC) is used as evaluation metrics for regression and classification task, respectively. 
	
	\subsection{Empirical Results}
	As shown in Table~\ref{exp_res}, our methods perform significantly better than other methods on both regression and classification task. For the school dataset, the averaged RMSE is reported. As we can see, our algorithm achieves a 15\% \textasciitilde 21\% improvement against the other MTL methods, and an 11\% \textasciitilde 24\% against GBT and IBT. One possible reason is that the sample size is small for each task, which is not enough to train a task-specific model, while task-specific information is ignored in the global model trained with all samples.
	
	In fraud detection tasks, the AUC for eash task is shown. We can find that the compared MTL methods perform pretty unsatisfactory. 
	Our algorithms perform much better than GBT and IBT, especially in the scenario where the sample sizes of different tasks are severely imbalanced, which means that the global model may be easily dominated by the task with bigger sample size. 
	%
	\begin{figure}[htbp]
		\centering
		\includegraphics[width=7cm, clip]{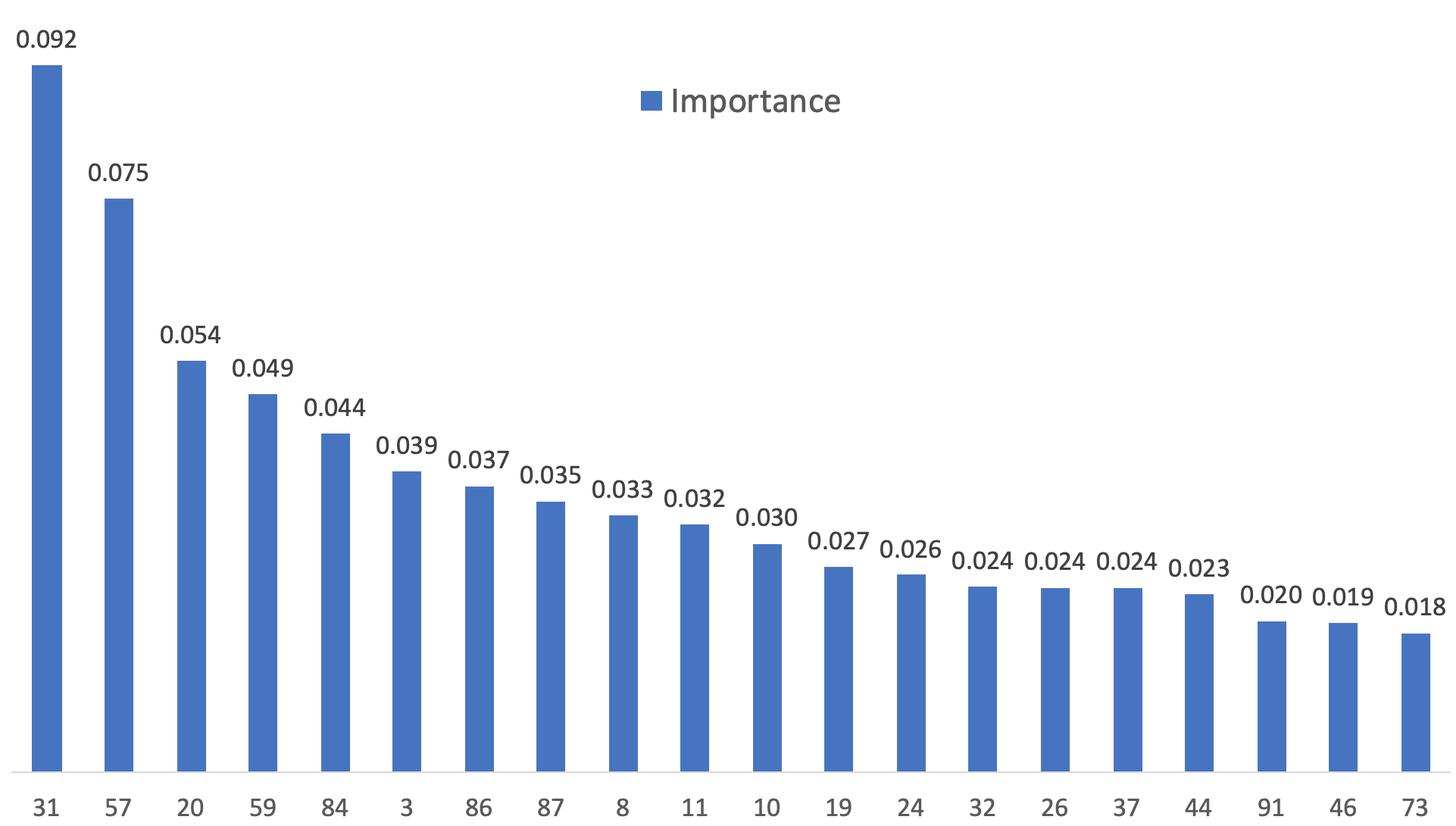}
		\caption{\small{The top 20 important features in task2 of Scene1.}}
		\label{fea_importance}
	\end{figure}

	\subsection{Interpretability}
	The interpretability of boosted tree model in both global and local level has been shown in~\cite{DBLP:conf/dasfaa/FangZLZ18}. In our work, since the whole model of each task consists of the common part and the specific part, so we collect them all to get the whole importance of each feature. For each instance, the contribution of each feature to the final prediction can be calculated with the method in~\cite{DBLP:conf/dasfaa/FangZLZ18}. An example of the top 20 important feature in task2 of Scene1 is shown in figure ~\ref{fea_importance}.
	
	\section{CONCLUSIONS}
	\label{con}
	This paper proposes a boosted tree based MTL method for heterogeneous tasks. 
	A two-stage method is developed with many strategies (i.e., regularization and early stopping) used.
	The experiments validate the effectiveness of the proposed method. 

	\bibliographystyle{ACM-Reference-Format}
	\bibliography{acmart}	
	
\end{document}